\documentclass[twocolumn]{article} 

\usepackage[american]{babel}
\usepackage{fullpage}
\usepackage{natbib} 
\usepackage{mathtools} 
\usepackage{booktabs} 
\usepackage{tikz} 

\usepackage{hyperref}
\usepackage{amsmath, amssymb, amsthm,bm}
\usepackage{algorithm}
\usepackage{algorithmic}
\usepackage{microtype}
\usepackage{graphicx}
\usepackage{subfigure}
\usepackage{booktabs} 

\usepackage{xcolor}
\usepackage{cleveref}
\usepackage{stfloats}
\graphicspath{{figs/}}

\theoremstyle{plain}
\newtheorem{theorem}{Theorem}[section]

\newtheorem{corollary}[theorem]{Corollary}
\newtheorem{assumption}[theorem]{Assumption}

\theoremstyle{definition}

\theoremstyle{remark}

\newcommand{\E}{\mathbb{E}}
\newcommand{\I}{\mathbf{\I}}

\newcommand{\G}{\mathbf{G}}
\newcommand{\R}{\mathbb{R}}
\renewcommand{\P}{\mathbf{P}}
\newcommand{\X}{\mathbf{X}}
\newcommand{\y}{\mathbf{y}}
\newcommand{\f}{\mathbf{f}}

\newcommand{\w}{\mathbf{w}}
\newcommand{\s}{\mathbf{s}}

\newcommand{\x}{\mathbf{x}}

\renewcommand{\f}{\mathbf{f}}

\newcommand{\bet}{\bm{\beta}}
\newcommand{\thet}{\bm{\theta}}

\newcommand{\Eps}{\bm{\epsilon}}

\def\R{\mathbb{R}}
\def\RF{\textup{RF}}

\usepackage[textsize=tiny]{todonotes}
\newcommand{\indict}{\mathbb{I}}

\title{Sparse Neural Additive Model: Interpretable Deep Learning with Feature Selection via Group Sparsity}
\author{Shiyun Xu\thanks{Department of Applied Mathematics and Computational Science, University of Pennsylvania. Email: {\tt shiyunxu@sas.upenn.edu}}
\and{Zhiqi Bu\thanks{Department of Applied Mathematics and Computational Science, University of Pennsylvania.}}
\and{Pratik Chaudhari\thanks{Department of Electrical and Systems Engineering, University of Pennsylvania.}}
\and{Ian J. Barnett\thanks{Department of Biostatistics, Epidemiology, and Informatics, University of Pennsylvania.}}}

\date{}

\begin{document}
\maketitle
\begin{abstract}
Interpretable machine learning has demonstrated impressive performance while preserving explainability. In particular, neural additive models (NAM) offer the interpretability to the black-box deep learning and achieve state-of-the-art accuracy among the large family of generalized additive models. In order to empower NAM with feature selection and improve the generalization, we propose the sparse neural additive models (SNAM) that employ the group sparsity regularization (e.g. Group LASSO), where each feature is learned by a sub-network whose trainable parameters are clustered as a group. We study the theoretical properties for SNAM with novel techniques to tackle the non-parametric truth, thus extending from classical sparse linear models such as the LASSO, which only works on the parametric truth. 

Specifically, we show that SNAM with subgradient and proximal gradient descents provably converges to zero training loss as $t\to\infty$, and that the estimation error of SNAM vanishes asymptotically as $n\to\infty$. We also prove that SNAM, similar to LASSO, can have exact support recovery, i.e. perfect feature selection, with appropriate regularization. Moreover, we show that the SNAM can generalize well and preserve the `identifiability', recovering each feature's effect. We validate our theories via extensive experiments and further testify to the good accuracy and efficiency of SNAM.
\end{abstract}

\section{Introduction}
Deep learning has shown dominating  performance on learning complex tasks, especially in high-stake domains such as finance, healthcare and criminal justice. However, most neural networks are not naturally as interpretable as decision trees or linear models. Even to answer fundamental questions like ``what is the exact effect on the output if we perturb the input?'', neural networks oftentimes rely on complicated and ad-hoc methods to explain the model behavior, with additional training steps and loose theoretical guarantee. As a result, the black-box nature of neural networks renders difficult and risky for human to trust deep learning models or at least to understand them.

There is a long line of work studying the interpretable machine learning. At high level, existing methods can be categorized into two classes: (1) model-agnostic methods, and (2) innately interpretable models. On one hand, model-agnostic methods aim to explain the predictions of models that are innately black-box, via the feature importance and local approximation, which include Shapley values \citep{shapley201617, vstrumbelj2014explaining, lundberg2017unified} and LIME \citep{ribeiro2016should} as the representatives. On the other hand, directly interpretable models such as the decision-tree-based models and the generalized additive models (GAM), including the generalized linear models (GLM, \citep{nelder1972generalized}) as sub-cases, are the most widely applied and demonstrate amazing performance.

To give more details, GLM is a powerful family of models that relates a linear model with its response variable by a link function $g$. 
\begin{align}
g(\E(\y))=\beta+\sum_{j=1}^{p} \beta_j \X_{ j}
\label{eq: GLM}
\end{align}

where $\y\in\R^n$ is the response and $\X_j$ is the $j$-th feature of the input matrix. However, such parametric form with $\beta_j$ limits the capacity of GLM when the unknown truth function takes a general and non-parametric form. This limitation motivates the development of GAM  \citep{hastie2017generalized}:
\begin{align}
g(\E(\y))=\beta+\sum_{j=1}^{p} f_{j}\left(\X_{ j}\right)\equiv \beta+f(\X). 
\label{eq: GAM}
\end{align}

Here $f_j$ is the unknown truth function (possibly non-linear) to be learned, which we refer to as the `effect'.

Recently, the neural additive model (NAM) \citep{agarwal2020neural} introduces a new member into the GAM family, which applies sub-networks to learn $f_j$ effectively, making accurate predictions while preserving the explainable power. Similar to regular neural networks, NAM learns a non-parametric model \eqref{eq: GAM} via its trainable parameters, instead of the functional approximation used by the traditional GAM. This parametric formulation allows NAM to be trained efficiently by off-the-shelf optimizers such as Adam. In addition, NAM can work flexibly with regression and classification problems, leveraging arbitrary network architecture to approximate $f_j$, hence fully exploiting the expressivity of deep learning.

\begin{figure}[!htb]
    \centering
    \includegraphics[width=1.05\linewidth,height=0.7\linewidth]{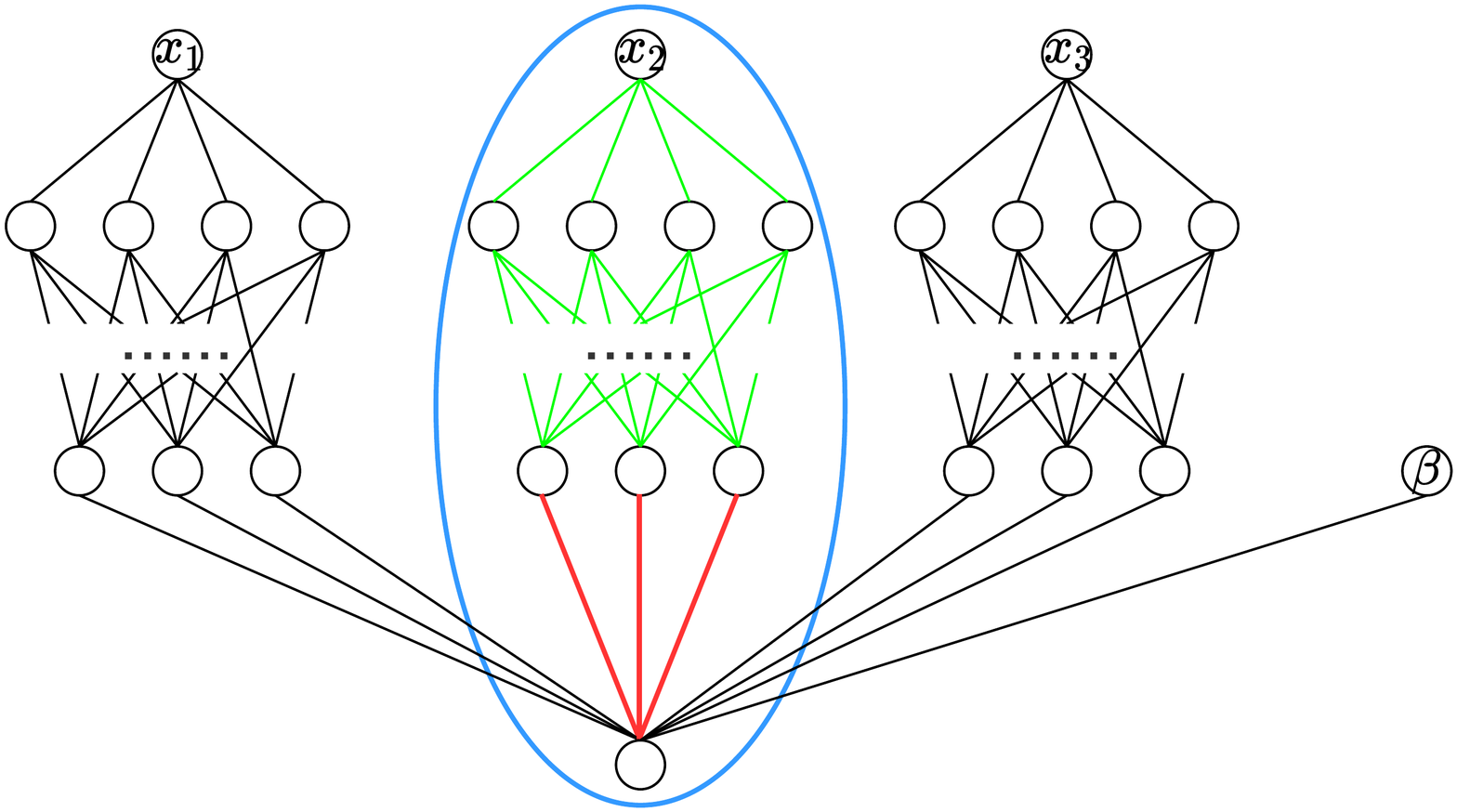}
    \hspace{-0.3cm}
    \vspace{-2cm}
    \caption{Architecture of NAM, with each sub-network (blue circle) being a group for Group LASSO regularization in SNAM. Note that in multi-class, multi-label, and multi-task problems, the last layer can have multiple neurons.}
    \label{fig:NAM architecture}
\end{figure}

Yet, theoretical results about NAM on some important questions are missing: Does the convergence of NAM behave nicely? Does NAM guarantee to learn the true additive model consistently, as sample size increases? How to modify NAM such as to select features and whether the feature selection is accurate? Can we expect each sub-network in NAM to recover each $f_j$? 

In this paper, we answer these questions in the affirmative. We study the sparse NAM with specific group sparsity regularization, especially the Group LASSO \citep{meier2008group,friedman2001elements}, which reduces to NAM when the penalty is zero. We highlight that our SNAM is the first innately interpretable model that simultaneously uses neural networks and allows feature selection. Our contributions are as follows:
\begin{enumerate}
\item We propose an innately interpretable model -- sparse neural additive model (SNAM) -- to empower NAM with feature selection. In particular, SNAM can employ the Group LASSO penalty that regularizes each sub-network's parameters as a group. Notice that LASSO is a special case of SNAM, when each sub-network has only one parameter. Our design easily extends to other SNAMs when we consider different group penalty such as the Group SLOPE.
\item We employ efficient and scalable optimizers, such as the subgradient and proximal methods (see \Cref{app:train SNAM}), to train SNAM. Consequently, we demonstrate its prediction power and trainability.
\item We establish an interesting connection between the LASSO and SNAM with Group LASSO regularization. Building on top of this, we rigorously derive the slow rate and the support recovery of SNAM. We show that SNAM approximates the true model, selects important features in a sample-efficient manner, and identifies individual functions $f_j$ asymptotically.
\item We empirically validate our theoretical results via synthetic and real datasets, further illustrating that SNAM is trainable, accurate, effective in feature selection, and capable of effect identification. For example, SNAM can be 3 times faster than SPAM (see \Cref{table:regression}) and save roughly half of parameters in NAM, while preserving comparable performance (see \Cref{table:compas}).
\end{enumerate}

For theoretical analysis, we focus on
\begin{align}
\y=\sum_{j=1}^{p} f_{j}\left(\X_{ j}\right)+\Eps
\label{eq:true model}
\end{align}
where i.i.d. samples $\X_j \sim \mathcal{X}_j$ for $j\in [p]$ where $\mathcal{X}_j$ is some distribution and the noise $\Eps\sim SG(\sigma^2)$ where $SG$ means sub-Gaussian with variance $\sigma^2$. For algorithms and experiments, we extend to GAM in \eqref{eq: GAM}.

\section{Additive Models in a Nutshell}
Linear regression is one of the most classic model, on which various extensions are based. One extension is the LASSO \citep{tibshirani1996regression}, a linear model that adds $\ell_1$ penalty to the linear model. This penalty not only empowers ordinary linear regression with feature selection but also regularizes the model against overfitting. Another extension is the GLM, which adds a link function to relate the linear model with its response to work on more general problems (e.g. logistic regression for classification). Note that GLM can combine with the $\ell_1$ penalty to give sparse logistic regression. 

While GLMs are all additive and thus directly interpretable, GAMs further improve the capacity of models by introducing the non-linearity, for instance, in NAM \citep{agarwal2020neural} and Explainable Boosting Machines (EBM) \citep{lou2012intelligible,nori2019interpretml}. In this work, we focus on NAM, a state-of-the-art GAM that incorporates neural networks and uses four types of regularization: dropout, weight decay ($\ell_2$ penalty), output penalty, and feature dropout. Unfortunately, all these types of regularization do not enable feature selection for NAM. 

Traditionally, one can only allow feature selection on GLMs (with $\ell_1$ regularization) or a few special GAMs, e.g. sparse additive model (SPAM by \citet{ravikumar2009sparse}, restated in \Cref{alg:SPAM}). As introduced in this paper, SNAM is a new member of GAM with feature selection. In addition, SNAM is the only GAM that is parametric (i.e. containing parameters that are trainable by gradient methods) besides GLMs: traditionally additive models are learned via the `backfitting algorithm'\footnote{The backfitting algorithm can be recovered from \Cref{alg:SPAM} when $\lambda=0$.} \citep{breiman1985estimating}, while neural networks are learned via gradient methods.

\begin{center}
\fbox{LASSO$\ \subseteq\ $GLM$\ \subseteq\ $NAM$\ \subseteq\ $SNAM$\ \subseteq\ $GAM}
\end{center}

One drawback of the backfitting algorithm is that the computation time will increase linearly with the number of features. This is due to the asynchronous or sequential estimation for each feature and a lack of theoretical understanding from the convergence viewpoint. The other drawback is the heavy memory complexity when executing the `smoothing' function (usually some smooth kernel splines) on large sample size. In fact, SNAM can out-speed SPAM by 3 times in \Cref{table:regression} on synthetic datasets, and SPAM runs out of memory on all real datasets considered here.

We give a brief summary of additive models in \Cref{tab:my_label}.

\begin{table}[!hbp]
    \centering
    \resizebox{!}{0.215\linewidth}{
    \begin{tabular}{|c|c|c|c|c|}
\hline
     Models&Non-linear&Non-param &Parametric&Feature
     \\&model&truth&model&selection
     \\
\hline
     LASSO&No&No&Yes&Yes \\
\hline
     GLM&No&No&Yes&Yes \\
\hline
     EBM&Yes&Yes&No&No \\
     (Trees)&&&&\\
\hline
     NAM&Yes&Yes&Yes&No \\
\hline
    SPAM&Yes&Yes&No&Yes \\
\hline
 SNAM&Yes&Yes&Yes&Yes \\
\hline
    \end{tabular}
    }
    \caption{Summary of additive (interpretable) models. In `Non-param truth', Yes/No means whether a model works without assuming that the truth is parametric.}
    \label{tab:my_label}
\end{table}

\section{SNAM: Model and Optimization}
\subsection{Model and Linearization Regimes}
To analyze SNAM under the regularization, for the $j$-th sub-network, we write the trainable parameters of  as $\bm\Theta_j$ (visualized in \Cref{fig:NAM architecture} by the blue circle) and the output as $h_j$. Then we write the SNAM output as
\begin{align*}
\begin{split}
h(\X,\bm\Theta)&={\sum}_{j} h_j(\X_j,\bm\Theta_j)+\beta
\end{split}
\end{align*}
With these notations in place, we can learn the model via the following SNAM optimization problem with some group sparsity regularization and an arbitrary loss $\mathcal{L}$:
\begin{align}
\min_{\bm\Theta,\beta}\mathcal{L}\big(\y,\sum_j h_j(\X_j,\bm\Theta_j)+\beta\big)+\text{GroupSparsity}(\{\bm\Theta_j\}).
\label{eq:general joint SNAM problem}
\end{align}

Notably, the group structure defined on sub-networks is the key to feature selection in SNAM: it explicitly penalizes $\bm\Theta_j$ so that the entries in $\bm\Theta_j$ are either all non-zero or all zero. The latter case happens when $\lambda$ is large, resulting in the $j$-th feature to be not selected as $h_j=0$.


In fact, if each sub-network has only a single parameter $\beta_j$ and no hidden layers at all, then the Group LASSO penalty is equivalent to the LASSO penalty: $\|\beta_j\|_2=|\beta_j|$. Therefore, we view LASSO as the simplest version of SNAM with Group LASSO regularization. This connection leads to the theoretical findings in this work, since we will analyze the linearization of SNAM. 

A long line of researches that linearizes the neural networks can be categorized into two main regimes: the neural tangent kernel (NTK) and the random feature (RF). The NTK regime linearizes the network under the `lazy training' constraint, where $\bm\Theta(t)\approx\bm\Theta(0)$ during entire training process, by applying a first-order Taylor expansion at $\bm\Theta(0)$. This lazy training phenomenon is usually guaranteed using the extremely (even infinitely) wide neural networks, and without any regularization\footnote{Unfortunately, $\bm\Theta(t)$ will be pushed away from its initialization $\bm\Theta(0)$ towards zero even under weak regularization, breaking the lazy training assumption \citep{fang2021mathematical,NEURIPS2020_9afe487d}.} \citep{jacot2018neural,xiao2020disentangling,arora2019exact,du2018gradient,allen2019convergence,bu2021dynamical,zou2020gradient}. Such limitation renders the NTK analysis invalid for SNAM.

The other branch of work uses the RF regime \citep{neal1996priors, rahimi2007random, yehudai2019power, ghorbani2021linearized} to linearize the neural network by fixing the weights in all hidden layers after initialization, and only training the output layer's weights. Mathematically, we decompose $\bm\Theta_j=[\w_j,\thet_j]$. We denote $\w_j$ as the weights of all hidden layers (green in \Cref{fig:NAM architecture}) and $\thet_j\in\R^m$ as the weights in the output layer (red in \Cref{fig:NAM architecture}). Then we can rewrite the output of SNAM as
\begin{align}
\begin{split}
h(\X,\w,\thet)&={\sum}_{j} h_j(\X_j,\w_j,\thet_j)+\beta
\\
&={\sum}_{j} g_j(\X_j,\w_j)\thet_j+\beta
\end{split}
\label{eq: SNAM model}
\end{align}
in which $\thet:=[\thet_1,\cdots,\thet_p], \w:=[\w_1,\cdots,\w_p]$, and the feature map $g_j:\R\to\R^m$ is the forward propagation of the $j$-th sub-network until the output layer. 

In this RF regime, SNAM is linear in trainable parameters $\thet$ (though non-linear in input $\X$) and is indeed a kernel regression, a topic with rich theoretical understanding.

\subsection{Group Sparsity and Optimization Problems}
It is well-known that group sparsity allows all parameters in the same group to be simultaneously non-zero or zero. One popular choice is the Group LASSO, with which the SNAM problem becomes
\begin{align}
\min_{\bm\Theta,\beta}\mathcal{L}\big(\y,\sum_j h_j(\X_j,\bm\Theta_j)+\beta\big)+\lambda\sum_{j}\|\bm\Theta_j\|_2.
\end{align}

For another example, we may consider the Group SLOPE:
\begin{align}
\min_{\bm\Theta,\beta}\mathcal{L}\big(\y,\sum_j h_j(\X_j,\bm\Theta_j)+\beta\big)+\sum_{j}\lambda_j\|\bm\Theta\|_{2,(j)},
\end{align}
where the penalty is a decreasing vector $(\lambda_1,\cdots,\lambda_p)$ and $\|\bm\Theta\|_{2(j)}$ denotes the $j$-th largest element in $\{\|\bm\Theta_1\|_2,\cdots,\|\bm\Theta_p\|_2\}$. We demonstrate other choises of group sparsity in \Cref{app:different SNAM}. In what follows, we focus on SNAM with the Group LASSO.

\subsection{Random Feature SNAM}
We study the RF neural network as a sub-class of SNAM, with two desirable benefits: (i) we do not restrict to weak (infinitesimal) regularization as in \cite{wei2019regularization}; (ii) we do not need neural networks to be wide. For the ease of presentation, we omit the output layer bias $\beta$:
$$h^{\RF}(\X,\thet)=\sum_{j=1}^{p}h_j^{\RF}(\X_j,\thet_j)=\sum_{j=1}^{p}\G_j\thet_j$$

where the random features $\G_j:=g_j(\X_j,\w(0))\in\R^{n\times m}$. Therefore, the corresponding optimization for the RF network is
\begin{align}
\hat\thet^{\RF}:=\text{argmin}_{\thet}\mathcal{L}(\y,\G\thet)+\lambda\sum_j\|\thet_j\|_2
\label{eq: general RF SNAM problem}
\end{align}

where $\G:=[\G_1,\cdots,\G_p]$ is the concatenation of $\G_j$.

\subsection{Convergence of SNAM and RF}
Algorithmically speaking, the general SNAM \eqref{eq:general joint SNAM problem} can be efficiently optimized by existing optimizers, e.g. the subgradient methods \citep{shor2012minimization,boyd2003subgradient} and the proximal gradient descent (ProxGD) \citep{nitanda2014stochastic,li2015accelerated,parikh2014proximal} (c.f. \Cref{app:train SNAM} for details), with possible acceleration (for example, subgradient Adam and Nesterov-accelerated ProxGD \citep{beck2009fast,su2014differential}). In fact, we can show that the subgradient descent and ProxGD both provably find the minimizer of SNAM \eqref{eq:general joint SNAM problem} and its RF variant \eqref{eq: general RF SNAM problem}. 

Denoting $\Theta$ to denote all trainable parameters in SNAM and $\Theta_j$ as those in the $j$-th sub-network, we claim both subgradient descent and ProxGD have the same gradient flow \citep[Section 4.2]{parikh2014proximal}:
\begin{equation*}
\frac{d \Theta}{d t}
=-\frac{\partial (\mathcal{L}(\y,h(\X, \Theta))+\lambda\sum_j\|\Theta_j\|_2)}{\partial \Theta}
\end{equation*}
Left multiply $\frac{\partial \Theta}{\partial t}^\top$ and integrate over time,
\begin{align*}
\int_0^\infty \left\|\frac{d \Theta}{d t}\right\|_2^2dt
&=\int_\infty^0 
\frac{d (\mathcal{L}(t)+\lambda\sum_j\|\Theta_j(t)\|_2)}{d t}dt
\\
&\leq \mathcal{L}(0)+ \lambda{\sum}_j \|\Theta_j(0)\|_2.
\end{align*}
Since the integral is increasing in time but upper bounded, we obtain that $\frac{d\Theta}{dt}\to 0$ and thus $\frac{d\mathcal{L}}{dt}\to 0$, i.e. $\mathcal{L}$ converges to the minimum. The convergence result implies the trainability of SNAMs (and NAMs as a by-product when $\lambda=0$) in practice. 
\begin{figure}[!htb]
    \centering
    \includegraphics[width=0.49\linewidth]{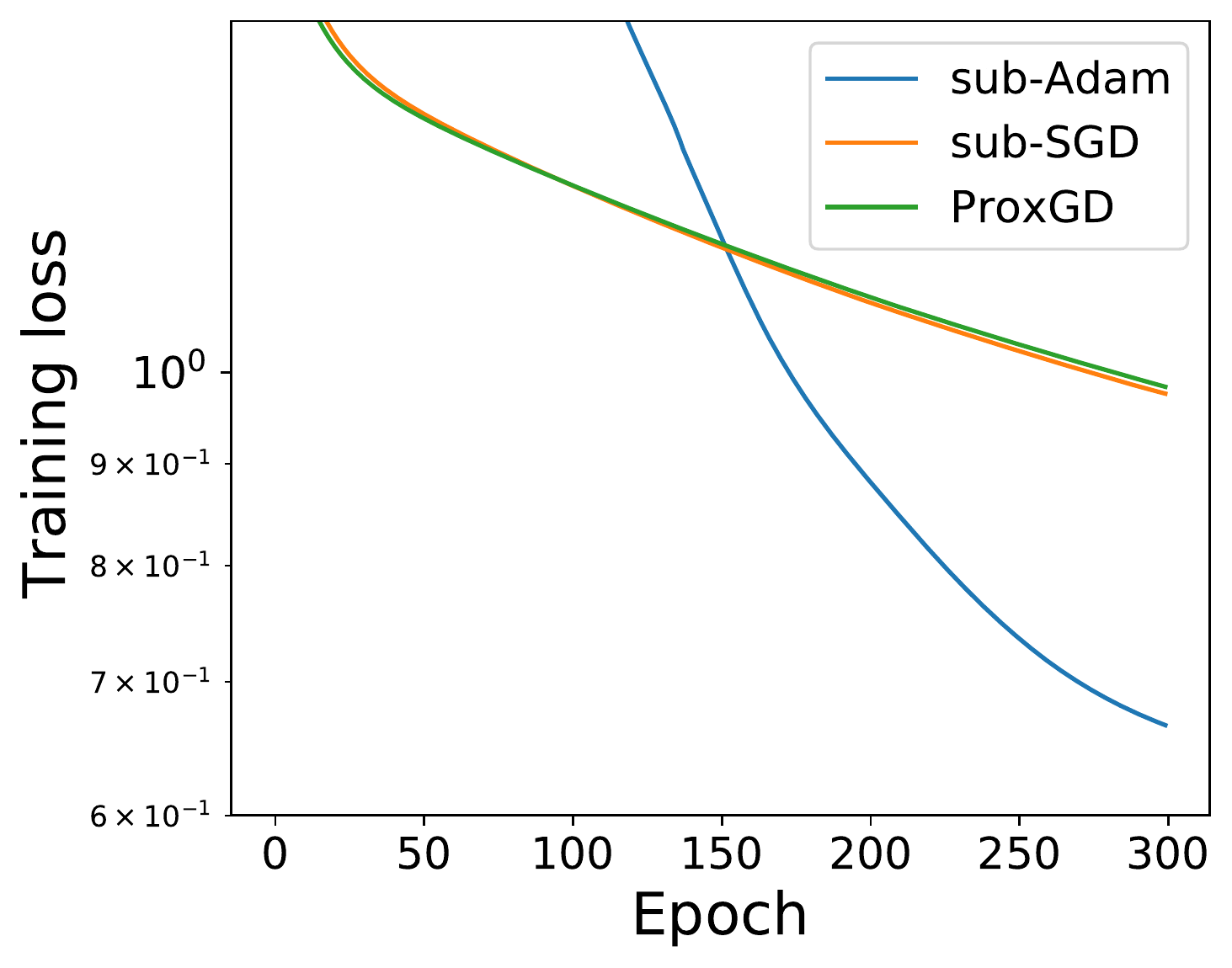}
    \includegraphics[width=0.49\linewidth]{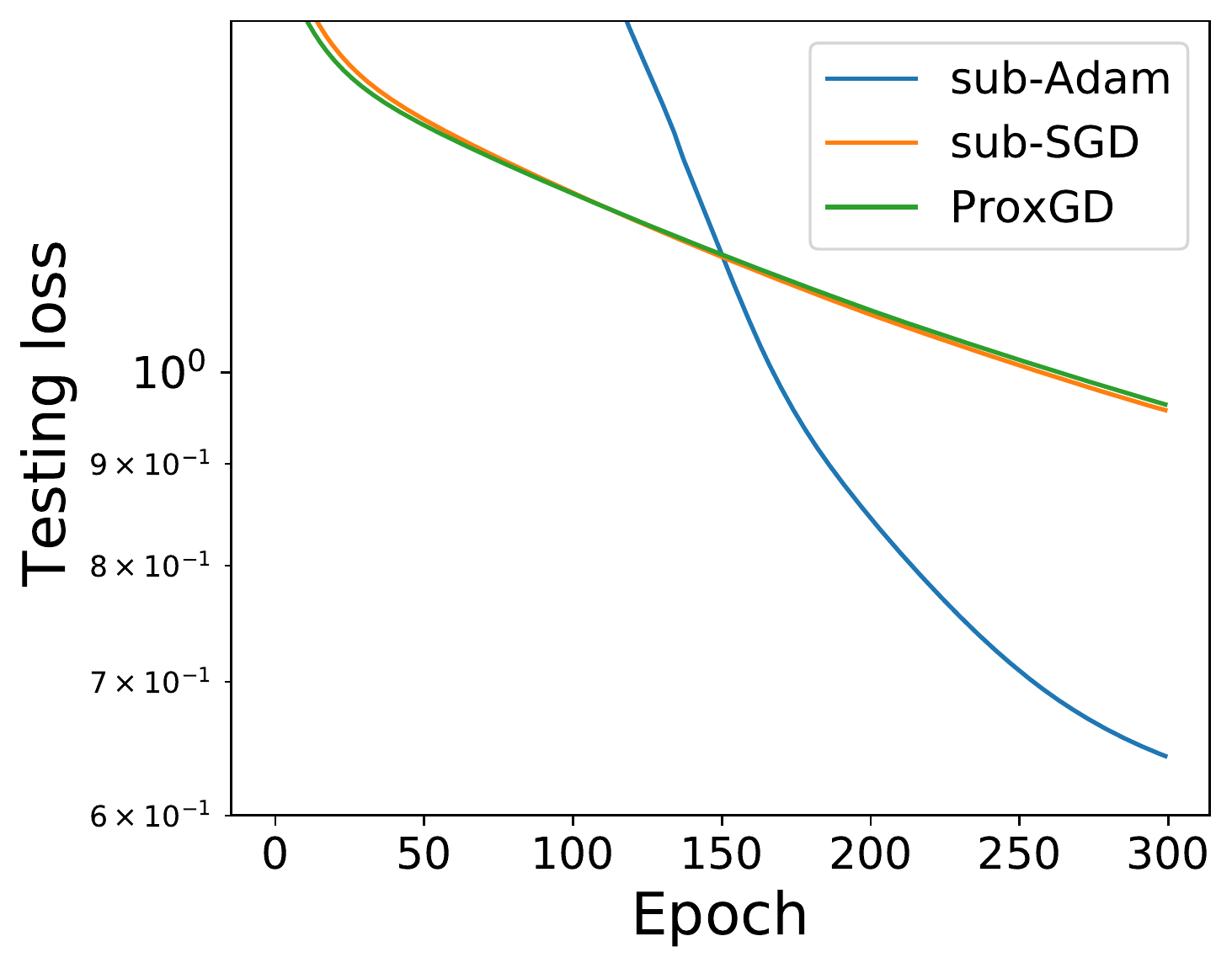}
    \caption{Loss convergence of SNAM on California Housing regression dataset, under different optimizers.}
    \label{fig:optimizer}
\end{figure}

Henceforth, we focus on the RF SNAM minimizer $\hat\thet^{\RF}$ in \eqref{eq:true model} and drop the super-script `$\RF$' for clearer presentation.

\section{Non-Asymptotic Analysis of SNAM}
In this section, we show that SNAM can approximate the truth model well on training set and achieve exact support recovery with finite number of samples.

We study the primal problem
\begin{align}
\hat\thet:=\text{argmin}_{\thet}\frac{1}{2}\|\y-{\sum}_j \G_j\thet_j\|_2^2+\lambda{\sum}_{j}\|\thet_j\|_2
\label{eq:SNAM problem}
\end{align}
and equivalently the dual problem
\begin{align}
\hat\thet:=\text{argmin}_{\thet:\sum_j\|\thet_j\|_2\leq \mu}\frac{1}{2}\|\y-{\sum}_j \G_j\thet_j\|_2^2
\label{eq:SNAM problem dual}
\end{align}
We point out that although the analysis of SNAM is similar to that of LASSO at high level, our analysis is technically more involved and requires novel tools, due to the fact that the true model \eqref{eq:true model} is non-parametric (unlike the LASSO whose true model is parametric).

\subsection{Slow Rate with Group LASSO Penalty}
Similar to the analysis of slow rate for the LASSO \citep{wainwright2009sharp}, our analysis needs SNAM to overfit the training data under the low-dimensional $\G$ regime.
\begin{assumption}[Overfitting of SNAM]\label{assume:overfitting}
Denoting the truth $\f_j:=f_j(\X_j)$, we assume there exists $\mu$ such that
\begin{align*}
\frac{1}{n}\|\y-\sum_j \G_j\hat\thet_j\|_2^2
&\leq \frac{1}{n}\|\y-\sum_j \f_j\|_2^2=\frac{1}{n}\|\bm\epsilon\|_2^2.
\end{align*}
\end{assumption}

To guarantee a unique solution of SNAM, we further assume that the SNAM feature map $\G$ has full rank.
\begin{assumption}[Full rank of feature map]\label{assume:full rank}
$\G\in\R^{n\times M}$ has full column rank $M$ and thus $\G^\top\G\in\R^{M\times M}$ is invertible.
\end{assumption}
Here $M$ is the sum of numbers of neurons at the last hidden layer of each sub-network\footnote{When all sub-networks have the same architecture, we write $M=mp$ where the last hidden layer width $m$. More generally, suppose the $j$-th sub-network has last hidde layer width $m_j$, then $M=\sum_j m_j$.}. Our first result is the slow rate of the SNAM convergence $h(\X,\hat\thet)\to f(\X)$ as $n\to\infty$. We highlight the definition of estimation error $\|f(\X)-h(\X,\hat\thet)\|^2/n$, which is different from the prediction error $\|\y-h(\X,\hat\thet)\|^2/n$.

\begin{theorem}\label{thm:slow_rate}
Under \Cref{assume:overfitting} and \Cref{assume:full rank}, supposing $|f_j|$ is upper bounded by constant $c_j$ and noise $\epsilon\sim SG(\sigma^2)$, then with probability at least $1-\delta_1-\delta_2$, we have for $\hat\theta$ in \eqref{eq:SNAM problem dual},
\begin{align*}
&\frac{1}{n}\|\sum_j (\f_j- \G_j\hat\thet_j)\|_2^2\leq \frac{2\sigma}{\sqrt{n}}\Bigg(\sum_j c_j/\sqrt{\delta_2}\\
&+ \mu\max_j\sqrt{\E g_j(\mathcal{X}_j,\w_j(0))^2}\sqrt{2\log( m_j/\delta_1)}\Bigg)
\end{align*}

where $m_j$ is the width of output layer in the $j$-th sub-network and $\mu$ is the penalty coefficient.
\end{theorem}
We refer the interested readers to \Cref{app:proofs} for the proof. In fact, we may further relax our assumption on the noise distribution in the true model \eqref{eq:true model}, at the cost of a strictly worse bound for any $\delta_1$.

\begin{corollary}
\label{cor:slow_rate}
Under \Cref{assume:overfitting} and \Cref{assume:full rank}, supposing $|f_j|$ is upper bounded by constant $c_j$ and noise has $\text{mean}(\epsilon)=0,\text{Var}(\epsilon)=\sigma^2$, then with probability at least $1-\delta_1-\delta_2$, we have for $\hat\theta$ in \eqref{eq:SNAM problem dual},
\begin{align*}
&\frac{1}{n}\|\sum_j (\f_j- \G_j\hat\thet_j)\|_2^2\leq \frac{2\sigma}{\sqrt{n}}\Bigg(\sum_j c_j/\sqrt{\delta_2}\\
&+\mu\max_j\sqrt{\E g_j(\mathcal{X}_j,\w_j(0))^2}\sqrt{ m_j/\delta_1}\Bigg)
\end{align*}
\end{corollary}
The proof only needs slight modification by leveraging the Kolmogorov inequality instead of the maximal sub-Gaussian inequality in \Cref{thm:slow_rate}. In both \Cref{thm:slow_rate} and \Cref{cor:slow_rate}, the MSE $\frac{1}{n}\|\sum_j (\f_j- \G_j\hat\thet_j)\|_2^2$ converges to zero with rate $1/\sqrt{n}$ as $n\to\infty$. We note that the convergence rate of SNAM has the same order as that of LASSO, but SNAM requires two probability quantities $\delta_1,\delta_2$ due to the non-parametric true model \eqref{eq:true model}, whereas the LASSO only needs $\delta_1$.

\subsection{Exact Support Recovery}
There has been a long line of researches on the support recovery, particularly on the parametric models such as the LASSO \citep{buhlmann2011statistics,wainwright2009sharp,tibshirani2017sparsity}, where the support is defined on the parameters, e.g. $supp(\bm{\hat\beta})=\{j:\hat\beta_j\neq 0\}, supp(\bm{\beta})=\{j:\beta_j\neq 0\}$, and the regularization is also defined on the parameters via $\lambda\|\hat\bet\|_1$. For non-parametric models like SPAM, the support is instead defined on the functions
$$S = supp(f)=\{j:f_j\neq 0\},$$
and the regularization is on the output function $\{h_j\}$. In contrast, our SNAM sets the sparse regularization on the parameters $\{\thet_j\}$, similar to LASSO. This explicit regularization allows us to borrow from the rich results of traditional support recovery for the LASSO and extend them to SNAM.

First, we assume that an insignificant feature ($j\not\in S$) is small when regressing on the true features.
\begin{assumption}[Mutual incoherence]\label{assume:mutual incoherence}
For some $ \gamma>0 $, we have
\begin{align}
 \Big\|\left(\G_{S}^\top \G_{S}\right)^{-1} \G_{S}^\top \G_{j}\Big\|_{2} \leq 1-\gamma, \text { for } j \notin S
\end{align}

where $\G_{S}$ is the concatenation of $\G_{j}$ for all $j\in S$.
\end{assumption}
Next, we assume that the regularization is not too large to omit significant features.
\begin{assumption}[Maximum regularization]\label{assume:regularization small}
The Group LASSO penalty coefficient $\lambda$ in \eqref{eq:SNAM problem} is small enough so that the following solution is dense 
\begin{align}
{\tilde\thet}_S:=\textnormal{argmin}_{\thet_S}\frac{1}{2}\|\y-\sum_{j\in S} \G_j\thet_j\|_2^2+\lambda\sum_{j\in S} \|\thet_j\|_2
\label{eq:a linear S}
\end{align}
\end{assumption}

We define the support of any prediction function $h(\cdot;\hat\thet)$ in two equivalent ways: one on the function and the other on the parameters,
$$supp(h)\equiv\{j:h_j\neq 0\}\equiv\{j:\|\hat\thet_j\|_2\neq 0\}.$$
We prove in \Cref{app:proofs} that, with proper Group LASSO regularization, the SNAM recovers the true $supp(f)$ exactly.

\begin{theorem}\label{thm:support}
Under \Cref{assume:full rank}, \Cref{assume:mutual incoherence} and \Cref{assume:regularization small}, then
$$\lambda>\max_{j\not\in S}\|\G_j^\top\|_\infty\|\y\|_\infty/\gamma$$ 

guarantees that the SNAM solution $\hat\thet$ in \eqref{eq:SNAM problem} has the exact support recovery, i.e. $supp(h)=supp(f)$.
\end{theorem}

\section{Asymptotic Analysis of SNAM}

In this section, we study the asymptotic consistency of SNAM and hence indicate its good generalization behavior. Our results build on top of the asymptotic zero loss between the ground truth and the prediction on training data, given by the slow rate in \Cref{thm:slow_rate}. The proofs can be found in \Cref{app:proofs}.

\subsection{Consistency}
We show in \Cref{thm:consistency in measure} that the SNAM $h_n$, when trained on $n$ samples, converges to the unknown true model $f$ in a probability measure. In other words, large amount of data promises that SNAM as a whole function can learn the truth. 

\begin{theorem}\label{thm:consistency in measure}
Under the assumptions in \Cref{thm:slow_rate}, we have the convergence in probability measure:
\begin{align*}
\lim _{{n\to \infty }}\rho (\{x\in \mathcal{X}:|f(x)-h_{n}(x)|\geq \varepsilon \})=0 
\end{align*}

for arbitrarily small $\epsilon>0$. Here $\rho$ is the probability measure of $\mathcal{X}$, the joint distribution of data $\X$. In words, the prediction function $h_n$ converges to the true model $f$. 
\label{thm:whole_consis_func}
\end{theorem}


\subsection{Effect Identifiability}
Another more difficult challenge in the generalized additive models is the identifiability of individual effects, in the sense that we want to have $h_j\to f_j$ for all $j\in[p]$. Notice that since the identifiability is a stronger property than the consistency, we need to assume more about the feature distribution $\mathcal{X}_j$. We show that SNAM is capable of identifying the effects in \Cref{cor:identifiability}.

\begin{theorem}[Effect Identifiability]
\label{cor:identifiability}
Assuming $h_n\to f$ in probability measure of $\mathcal{X}$ as $n\to\infty$, if $\mathcal{X}_j$ is independent of $\mathcal{X}_{-j}$, then $\lim_{n\to\infty} h_{n,j}(x)$ converges to $f_j(x)$ in probability up to a constant.
\end{theorem}


\section{Experiments}
In this section, we conduct multiple experiments on both synthetic and real datasets. we emphasize that here SNAM is not RF SNAM, i.e. we train all parameters in sub-networks. All experiments are conducted with one Tesla P100 GPU. We use MSE loss for regression, cross-entropy (CE) loss for classification, and wall-clock time for all tasks. Furthermore, we compare SNAM to other possibly sparse interpretable methods: NAM, $\ell_1$ linear support vector machine (SVM), LASSO and SPAM \citep{ravikumar2009sparse}. Experiment details such as data pre-processing, model architecture and hyperparamters are listed in \Cref{app:experiment detail}.

\subsection{Synthetic Datasets}
To validate our statistical analysis on SNAM, i.e. the feature selection (or support recovery), the estimation consistency and the effect identifiability, we experiment on synthetic regression and classification datasets. We emphasize that, it is necessary to work with synthetic data instead of real-world ones, since we need access to the truth $f_j$ for our performance measures.

\subsubsection{Data generation}
We generate a data matrix $\X\in \R^{3000\times 24}$ and denote the $j$-th column of $\X$ as $\X_j$. $\y$ is generated by the following additive model, for regression and binary classification, respectively:
\begin{align*}
\y&=f_1(\X_1)+\cdots+f_{24}(\X_{24})+\mathcal{N}(0,1),
\\
\mathbb{P}(\y=1)&=\text{sigmoid}(f_1(\X_1)+\cdots+f_{24}(\X_{24})).
\end{align*}
where all $f_j$ are zero functions except
\begin{align*}
    f_1(\X_1) &= 2x^2\tanh{x}\\
    f_2(\X_2) &= \sin{x}\cos{x}+x^2\\
    f_3(\X_3) &= 20/(1+e^{-5\sin{x}})\\
    f_4(\X_4) &= 20\sin^3{2x}-6\cos{x}+x^2
\end{align*}

\subsubsection{Performance measures}
Denote the output of each sub-network as $\hat f_j$. To illustrate the performance on the support recovery, we use precision and recall to compare $\hat f_j$ and truth $f_j$. In particular, we use $\ell_2$ norm of a sub-network's weights to indicate whether $\hat f_j=0$.

We now introduce the identification error (iden. error),
\begin{align*}
&\min_{c_j\in \R}\frac{1}{n}\| \hat f_j(\X_{j})-f_j(\X_{j})-c_j\|_2^2
\\
=&\frac{1}{n}\| \hat f_j(\X_{j})-f_j(\X_{j})-\hat c_j\|_2^2
\end{align*}
in which $\hat c_j := \frac{1}{n}\sum_{i=1}^n
(\hat f_j(\X_{ij})-f_j(\X_{ij}))$. Notice that \Cref{cor:identifiability} claims the convergence up to a constant $\hat c_j$.

\subsubsection{Results}
\begin{table}[!ht]
	\centering
	\begin{tabular}[t]{ccccc}
		\hline
		&$\ell_1$ SVM&LASSO&SPAM &SNAM\\
		\hline
		MSE loss&140.7&139.7&25.75&\textbf{10.61}\\
		Precision &0.17&\textbf{1.00}&0.17&\textbf{1.00}\\
		Recall &\textbf{1.00}&\textbf{1.00}&\textbf{1.00}&\textbf{1.00}\\
		Iden. error&5.90&6.09&3.07&\textbf{0.69}\\
		Time (sec) &0.005&0.007&152.1&48.52\\
        \#. Feature &24&4&4&4\\
        \#. Param &24&4&-&127201\\
		\hline
	\end{tabular}
    \caption{Performance of sparse interpretable methods on synthetic regression.}
	\label{table:regression}
\end{table}

\begin{figure}[!htb]
    \centering
    \includegraphics[width=\linewidth]{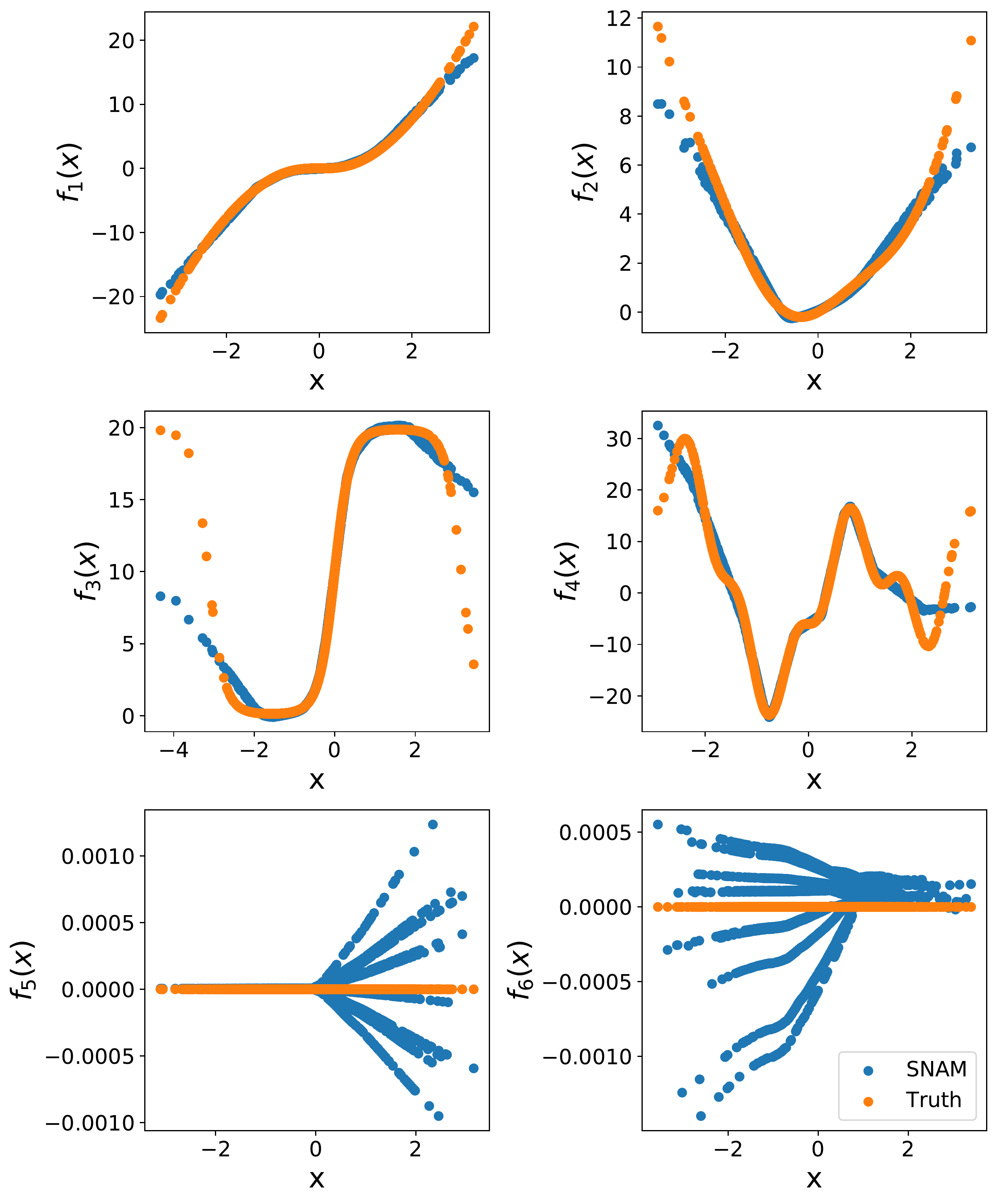}
    \caption{Individual effect learned by SNAM on synthetic regression. Blue dots are prediction $\hat f_j(\X_j)$ and orange dots are truth $f_j(\X_j)$, with $j=1,\cdots,6$.}
    \label{fig:regression_snam}
\end{figure}

In \Cref{table:regression}, for regression task, SNAM domintes existing sparse interpretable methods in all measures. Especially, SNAM (which includes LASSO as a sub-case) is the only method that achieves exact support recovery, obtaining perfect precision and recall scores. When facing complicated target functions, SNAM, as a non-linear model, significantly outperforms linear models like linear SVM and LASSO, in terms of test loss and identification error. In contrast to SPAM, another non-linear model that achieves low loss, SNAM outperforms in both loss and efficiency, with a 3 times speed-up. We further visualize the effects learned by SNAM in \Cref{fig:regression_snam}, demonstrating the strong approximation offered by the neural networks, and leave those learned by other interpretable methods in \Cref{app:fig zoo}.

\begin{table}[!htb]
	\centering
	\begin{tabular}[t]{ccccc}
		\hline
		&$\ell_1$ SVM&LASSO&SPAM &SNAM\\
		\hline
		CE loss&0.27&0.26&-&\textbf{0.15}\\
		Test accuracy &73.2&74.2&-&\textbf{94.1}\\
	Precision&0.57&0.67&-&\textbf{1.00}\\
	Recall&\textbf{1.00}&\textbf{1.00}&-&\textbf{1.00}\\
		Time (sec)&0.005&0.019&-&10.10\\
        \#. Feature &13&6&-&4\\
		\#. Param &13&6&-&128402\\
\hline
	\end{tabular}
    \caption{Performance of sparse interpretable methods on synthetic classification.}
    \label{table:classification}
\end{table}

Similarly in \Cref{table:classification}, for classification task, SNAM again significantly outperforms existing sparse interpretable methods: roughly 20\% higher accuracy and 33\% higher precision. Here LASSO means $\ell_1$ regularized logistic regression and SPAM cannot perform the classification in original text \citep{ravikumar2009sparse}.

\subsection{California Housing Regression}
California Housing \citep{pace1997sparse} is a dataset for studying the effect of community characteristics on housing prices in California districts from 1990 U.S. census. The task is to predict the median housing price based on 20640 examples and 8 features. In \cite{agarwal2020neural}, a well-trained NAM  deems the median income, latitude and longitude as the most significant features for an accurate prediction. Reassuringly, our SNAM concurs with the their conclusion by selecting the same features. Although the conclusion is the same, we highlight a key difference between the approaches: while the authors in \cite{agarwal2020neural} base their conclusion on the ad-hoc visual examination of the shape function $\hat f_j$, our approach is based on a hypothesis testing: $\thet_j=0$ v.s. $\thet_j\neq 0$ where $\thet_j$ is all parameters in a sub-network. We recognize a small decrease in the loss as the cost of feature selection, when compared to NAM, but SNAM can save 12.5\% in the number of parameters (or memory). Additionally, SNAM still outperforms other sparse interpretable methods. In fact, although SNAM takes longer to achieve its optimal performance in \Cref{tab:california}, it only takes about 14 seconds to outperform the optimal LASSO and SVM.

\begin{figure}[!htb]
    \centering
    \includegraphics[width=\linewidth]{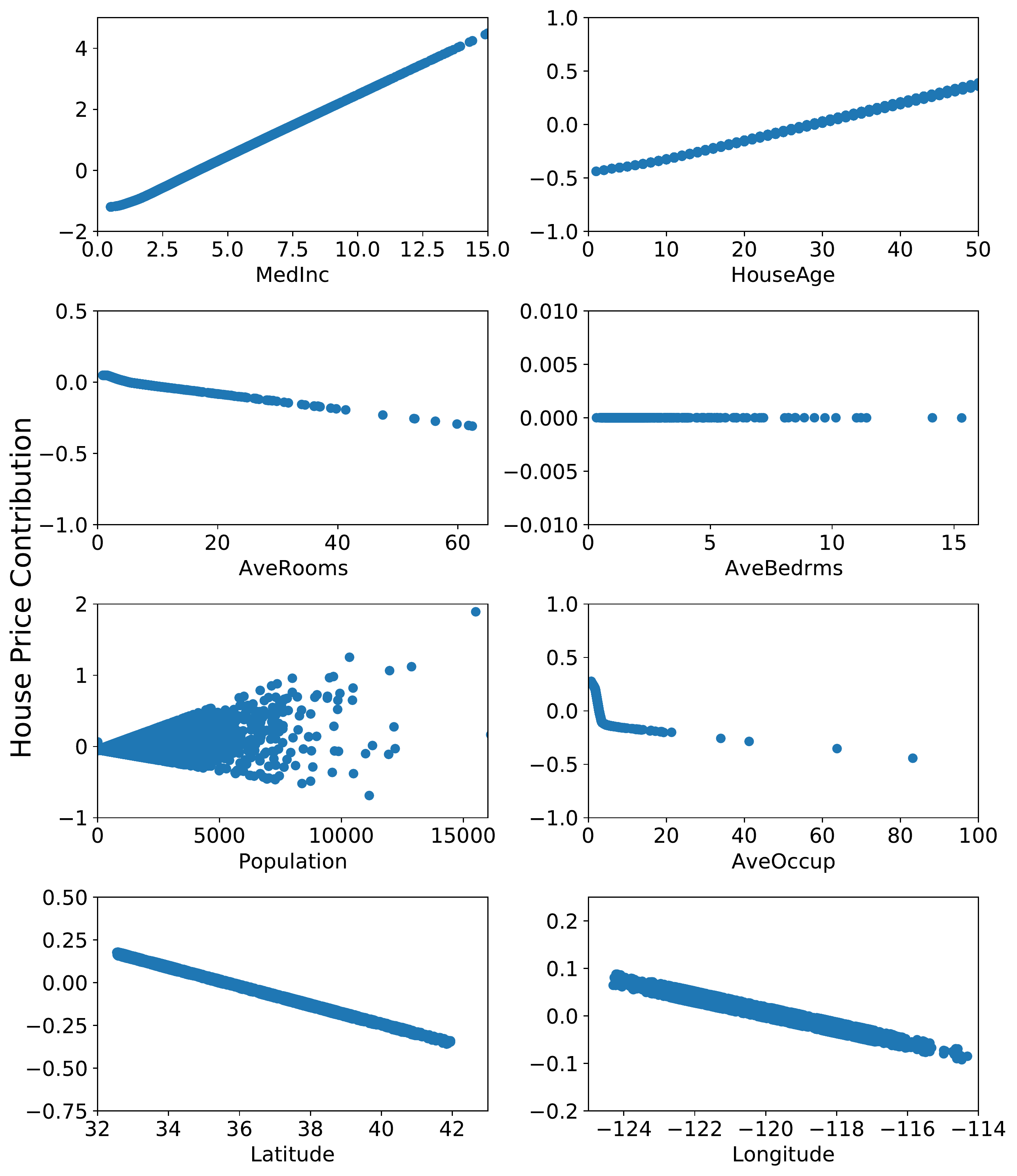}
    \caption{Individual effect learned by SNAM on California Housing dataset.}
    \label{fig:california}
\end{figure}

\begin{table}[!ht]
	\centering
	\begin{tabular}[t]{ccccc}
		\hline
		&$\ell_1$ SVM&LASSO&NAM &SNAM\\
		\hline
		MSE loss&0.654&0.712&\textbf{0.451}&\textbf{0.567}\\
		MAE loss&0.594&0.654&\textbf{0.479}&\textbf{0.526}\\
		$R^2$ score &0.501&0.457&\textbf{0.696}&\textbf{0.645}\\
		Time (sec)&1.37&0.01&343&340\\
        \#. Feature &6&2&8&7\\
\#. Param &6&2&42401&37101\\
		\hline
		\end{tabular}
		\label{tab:california}
    \caption{Performance of interpretable methods on California Housing dataset.}
\end{table}

\subsection{COMPAS Classification}
COMPAS is a widely used commercial tool to predict the recidivism risk based on defendants' features and it is known for its racial bias against the black defendants. The ProPublica released the recidivism dataset \citep{angwin2016machine}, that includes the characteristics of defendants in Broward County, Florida, and the predictions on reoffending by the COMPAS algorithm. This dataset has 6172 examples and 13 features \footnote{The data preprocessing follows \url{https://github.com/propublica/compas-analysis}.}.
\begin{table}[!ht]
	\centering
	\begin{tabular}[t]{ccccc}
		\hline
		&$\ell_1$ SVM&LASSO&NAM &SNAM\\
		\hline
		CE loss&\textbf{0.486}&\textbf{0.484}&0.503&0.504\\
		Test accuracy &75.3&75.4&75.3&\textbf{75.6}\\
		AUC score &\textbf{0.744}&\textbf{0.743}&0.714&\textbf{0.745}\\
		Time (sec)&0.106&0.175&27.5&27.4\\
\#. Feature &13&12&13&5\\
\#. Param &13&12&69552&26750\\
		\hline
	\end{tabular}
    \caption{Performance of interpretable methods on COMPAS dataset.}
    \label{table:compas}
\end{table}


\begin{figure}[!htp]
    \centering
    \includegraphics[width=0.8\linewidth]{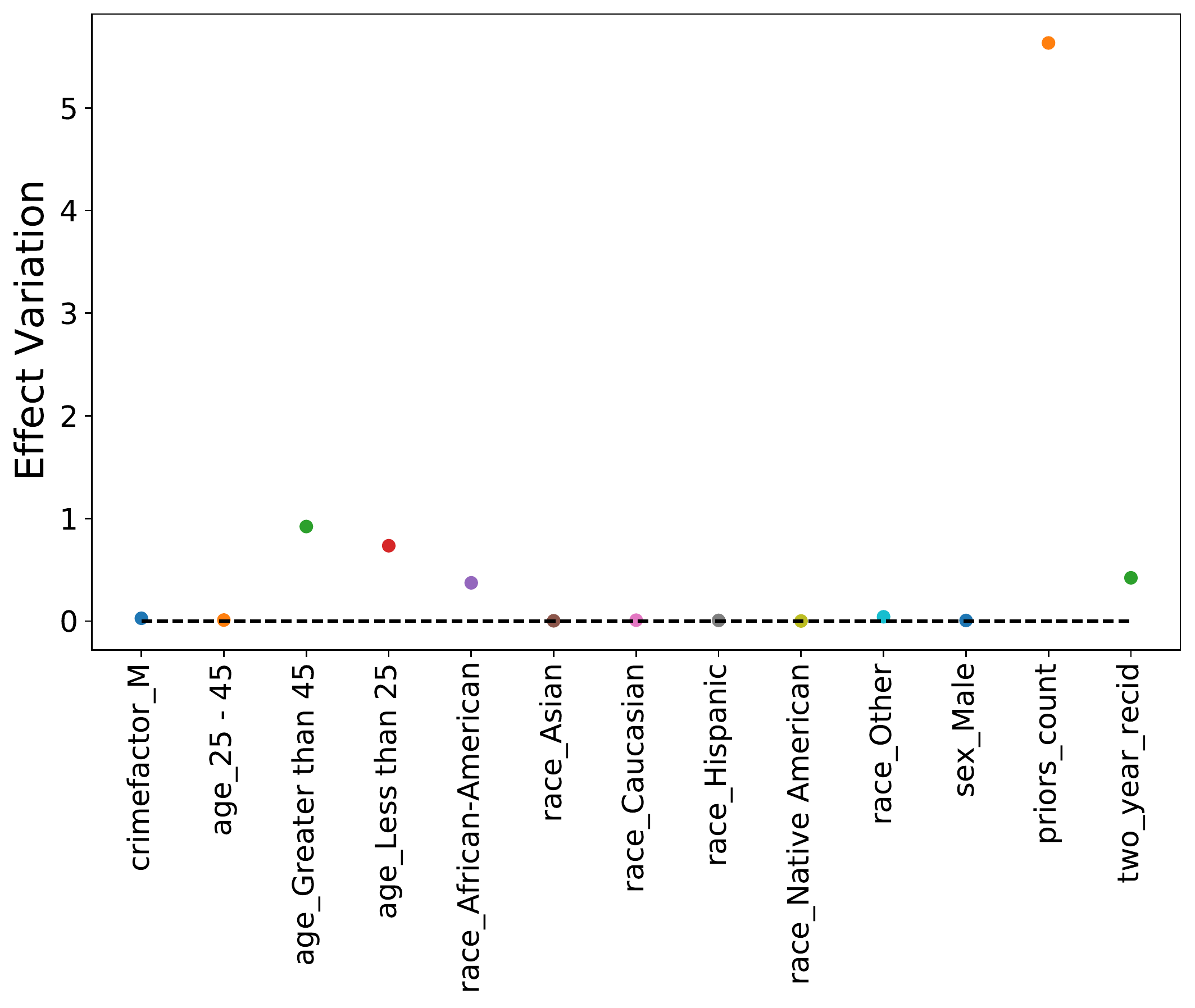}
    \caption{Variation of effects learned by SNAM on COMPAS dataset.}
    \label{fig:compas}
\end{figure}

In \Cref{table:compas}, we notice that all interpretable methods perform similarly, and SNAM has the highest AUC score between label and prediction, even though it only contains 54\% of NAM's parameters. A closer look at \Cref{fig:compas} describes the relations between features and the variation of effect, which is gap between the minimum recidivism risk and the maximum one among all individual samples for a particular feature, i.e. $\max_i\hat f_j(X_{ij})-\min_i\hat f_j(X_{ij})$. If the variation of an effect is large, then SNAM indicates the feature is significant. Indeed, the top 5 features selected by SNAM are prior counts, ages, two year recidivism and whether the defendant is African American. The last feature clearly demonstrates SNAM's explanability of the COMPAS algorithm's racial bias. In short, the features selected by SNAM are consistent with NAM's selection based on shape functions (c.f. \Cref{app:fig zoo}).

\subsection{Super-conductivity Regression}
We further experiment on the super-conductivity dataset from UCI repository, aiming to predict the critical temperature of super-conductors based on physical quantities (e.g. atomic radius, mass, density...) and chemical formulae. We highlight that the Super-conductivity is a high-dimensional dataset with 21263 samples and 131 features\footnote{The original dataset has 168 features. We remove the column \texttt{material} and all columns with variance less than 5\%.}, whereas all datasets in \cite{agarwal2020neural} have at most 30 features.

\begin{table}[!htb]
	\centering
	\begin{tabular}[t]{ccccc}
		\hline
		&$\ell_1$ SVM&LASSO&NAM &SNAM\\
		\hline
		MSE loss&410.0&311.7&\textbf{274.1}&\textbf{280.3}\\
		MAE loss&15.47&13.43&\textbf{11.56}&\textbf{12.09}\\
		$R^2$ score loss&0.654&0.731&\textbf{0.787}&\textbf{0.775}\\
		Time (sec)&5.00&1.87&682&688\\
        \#. Feature &100&50&131&72\\
        \#. Param &100&50&6289&3457\\
		\hline
	\end{tabular}
    \caption{Performance of interpretable methods on super-conductivity dataset.}
    \label{tab:super-conductivity}
\end{table}

We note that SNAM obtains similar performance as NAM and LASSO. In addition, the sparsity in SNAM saves 45\% number of parameters. In fact, given that NAM gives the best performance, a practitioner can always choose small penalty in SNAM in order to trade model efficiency for better performance.

\section{Discussion}
In this work, we propose the sparse neural additive model (SNAM) which applies a specific Group LASSO regularization explicitly to NAM. On one hand, SNAM is an interpretable deep learning model where the effect of each feature on the output can be extracted. On the other hand, the Group LASSO regularization empowers the network to select informative features, in the same way that LASSO empowers the linear model. We develop theoretical analysis of the optimization, the slow rate, the support recovery, the consistency of prediction, and the effect identifiability. Additionally, our experiments demonstrate the advantage of SNAM in memory and training efficiency, especially over non-regularized NAM and existing regularized interpretable methods. However, the superiority in performance usually comes at the price of longer training time than simpler methods like LASSO.

For future directions, one may further extend SNAM's theory to the fast convergence rate \citep{van2009conditions} in sample size, or to the jointly trained SNAM in terms of time. We believe the theoretical analysis and empirical evaluation can be explored for a whole family of interesting SNAMs. For example, while SNAM with Group LASSO penalty contains LASSO as sub-case, we can view SNAM with Group SLOPE \citep{brzyski2019group} penalty as extension of SLOPE \citep{bogdan2015slope}. Other possible extensions of elastic net \citep{zou2005regularization}, adaptive LASSO \citep{zou2006adaptive}, $K$-level SLOPE \citep{zhang2021efficient,bu2021characterizing} are also possible with SNAM (see \Cref{app:different SNAM} for examples).

\bibliography{uai_ref}

\clearpage
\onecolumn
\appendix
\providecommand{\upGamma}{\Gamma}
\providecommand{\uppi}{\pi}


%
%





%

%


\section{Proofs of Main Results}\label{app:proofs}

\subsection{Proof of Theorem 4.3}
\label{app:proof1}
\begin{proof}
By the Lagrange duality, 
for any penalty $\lambda>0$, there exists some $\mu>0$ such that the optimization problem
\begin{align*}
\min_{\thet} \frac{1}{2}\|\y-\sum_j \G_j\thet_j\|_2^2+\lambda\sum_j \|\thet_j\|_2
\end{align*}
is equivalent to 
\begin{align*}
&\min_{\thet} \frac{1}{2}\|\y-\sum_j \G_j\thet_j\|_2^2\text{\quad s.t. } \sum_j\|\thet_j\|_2\leq \mu 
\end{align*}
From \Cref{assume:overfitting}, the minimizer $\hat\thet$ satisfies that 
\begin{align}
&\frac{1}{n}\|\Eps+\sum_j (\f_j- \G_j\hat\thet_j)\|_2^2
=\frac{1}{n}\|\y-\sum_j \G_j\hat\thet_j\|_2^2\leq \frac{1}{n}\|\y-\sum_j \f_j\|_2^2=\frac{1}{n}\|\Eps\|_2^2.
\label{eq:eps}
\end{align}
Expanding the left-most term, 
\begin{align*}
&\frac{1}{n}\|\Eps+\sum_j (\f_j- \G_j\hat\thet_j)\|_2^2
= \frac{1}{n}\|\Eps\|_2^2+\frac{1}{n}\|\sum_j (\f_j- \G_j\hat\thet_j)\|_2^2
+\frac{2}{n}\left\langle \Eps, \sum_j (\f_j- \G_j\hat\thet_j) \right\rangle
\end{align*}
Substituting back to \eqref{eq:eps} and after some rearranging, we get:
\begin{align*}
&\frac{1}{n}\|\sum_j (\f_j- \G_j\hat\thet_j)\|_2^2
\leq \frac{2}{n}\sum_j\left\langle \Eps,  \G_j\hat\thet_j-\f_j \right\rangle\\
&\leq \frac{2}{n}\sum_j\left|\Eps^\top  (\G_j\hat\thet_j-\f_j) \right|\\
&\leq \frac{2}{n}\sum_j( |\Eps^\top\G_j\hat\thet_j|+|\Eps^\top \f_j|)\\
&\leq \frac{2}{n}\sum_j( \|\Eps^\top\G_j\hat\thet_j\|_2+\|\Eps^\top \f_j\|_2)\\
&\leq \frac{2}{n}(\sum_j \|\G_j^\top\Eps\|_{\infty}\|\hat\thet_j\|_2+\sum_j \|\f_j\|_\infty \|\Eps\|_2)\\
&\leq \frac{2}{n}(\sum_j \|\G_j^\top\Eps\|_{\infty}\|\hat\thet_j\|_2+\sum_j c_j \|\Eps\|_2)
\end{align*}
where the third inequality follows by the triangular inequality and the second last inequality holds by the Holder's inequality.
Note that $\|\G_j^\top\Eps\|_{\infty}=\max_{k=1,2,\cdots, m}|(\G_j^\top)_k\Eps|$ is a maximum of $m$ Gaussians. Here $(\G_j^\top)_k\in\R^n$ is the $k$-th feature fed into the output layer of the $j$-th sub-network. For each $k$, $(\G_j^\top)_k\Eps$ has mean zero and variance 
$$\text{Var}((\G_j^\top)_k\Eps)=\sigma^2\E((\G_j)_k^\top (\G_j)_k)=n\sigma^2 \E g_j(\mathcal{X}_j,\w_j(0))^2$$
By the maximal sub-Gaussian inequality \cite{boucheron2013concentration}, for any $\delta_1>0$, with probability at least $1-\delta_1$:
\begin{align*}
&\|\G_j^\top\Eps\|_{\infty}=\max_{k=1,2,\cdots, m}|(\G_j)_k\Eps|
\leq \sigma\sqrt{n\E g_j(\mathcal{X}_j,\w_j(0))^2}\sqrt{2\log(m_j/\delta_1)}.
\end{align*}

Furthermore, by Markov's inequality,
with probability at least $1-\delta_2$, we have $\|\Eps\|_2^2\leq \E(\|\Eps\|_2^2)/\delta_2= n\sigma^2/\delta_2$.
In summary, we obtain 
\begin{align*}
&\frac{1}{n}\|\sum_j (\f_j- \G_j\hat\thet_j)\|_2^2\\
&\leq \frac{2}{n}(\sum_j \|\G_j^\top\Eps\|_{\infty}\|\hat\thet_j\|_2+\sum_j c_j \|\Eps\|_2)\\
&\leq \frac{2}{\sqrt{n}}(\sum_j \sigma\sqrt{\E g_j(\mathcal{X}_j,\w_j(0))^2}\sqrt{2\log(m_j/\delta_1)}\|\hat\thet_j\|_2
+\sum_j c_j \sigma/\sqrt{\delta_2})\\
&\leq \frac{2\sigma}{\sqrt{n}}(\mu\max_j\sqrt{\E g_j(\mathcal{X}_j,\w_j(0))^2}\sqrt{2\log( m_j/\delta_1)}
+\sum_j c_j/\sqrt{\delta_2})
\end{align*}
\end{proof}

\subsection{Proof of Theorem 4.7}
\label{app:proof2}
For ease of presentation, we assume each sub-network has the same architecture, with last layer width $m$.
\begin{proof}
We construct and study a specific vector $\tilde\thet\in\R^{|S|m\times 1}$ by setting ${\tilde\thet}_S$ as in \eqref{eq:a linear S} and $\tilde\thet_j=\bm 0$ for $j\not\in S$: denoting the complement set of $S$ as $S^C$), we have:
\begin{align*}
&{\tilde\thet}_S=\textnormal{argmin}_{\thet_S}\frac{1}{2}\|\y-\sum_{j\in S} \G_j\thet_j\|_2^2+\lambda\sum_{j\in S} \|\thet_j\|_2 
\text{\quad and \quad} \tilde\thet_{S^C}=\bm 0.
\end{align*}
From \Cref{assume:regularization small} (maximum regularization), we have that $\tilde\thet_S$ is dense, i.e. $\tilde\thet_j\neq\bm 0$ for all $j\in S$. Therefore, if the constructed $\tilde\thet$ is indeed the SNAM solution $\hat\thet$ in \eqref{eq:SNAM problem}, then $supp(h)\supseteq supp(f)$. Further, $\tilde\thet_{S^C}=\bm 0$ leads to $supp(h)=S=supp(f)$.

Next, we check that the constructed $\tilde\thet$ is indeed the solution of SNAM in \eqref{eq:SNAM problem} via the KKT condition, which requires that for all $j\in[p]$,

\begin{equation}
\begin{aligned}
&\G_{j}^\top(\sum\nolimits_{l=1}^p {\G_{l}}\tilde\thet_l-\y)+\lambda \s_j
=\G_{j}^\top(\G_S\tilde\thet_S-\y)+\lambda \s_j=0
\label{eq:dL0}
\end{aligned}
\end{equation}
Here $\s_j$ is the subgradient of $\|\tilde\thet_j\|_2$, which is $\tilde\thet_j/\|\tilde\thet_j\|_2$ if $\tilde\thet_j\neq\bm 0$ and otherwise within a unit sphere. The first equality of \eqref{eq:dL0} follows by the construction $\tilde\thet_{S^C}=\bm 0$.
We break \eqref{eq:dL0} into the support set $S$ and its complement $S^C$,
\begin{align}
\G_{S}^\top\left(\y-\G_{S}\tilde\thet_S\right)&=\lambda \s_S
\label{eq:dL21}
\\
\G_{S^C}^\top\left(\y-\G_{S}\tilde\thet_S\right)&=\lambda \s_{S^C}
\label{eq:dL22}
\end{align}
Notice that if both KKT conditions \eqref{eq:dL21} and \eqref{eq:dL22} are satisfied by $\tilde\thet$, then $\tilde\thet=\hat\thet$. For $j\in S$, the KKT condition in \eqref{eq:dL21} is the same as that of \eqref{eq:a linear S} and hence satisfied by the definition of $\tilde\thet_S$. For $j\not\in S$, our goal is to show $\|\s_j\|_2<1$, which is a sufficient condition to guarantee $\tilde\thet_{S^C}=\bm 0$ and thus to satisfy the KKT condition \eqref{eq:dL22}. 

To show $\|\s_j\|_2<1$, we can solve $\tilde\thet_S$ from \eqref{eq:dL21}, leveraging the full rank of $\G_{S}^\top\G_{S}\in\R^{|S|m\times |S|m}$ from \Cref{assume:full rank}, and obtain
$$\tilde\thet_S=\left(\G_{S}^\top\G_{S}\right)^{-1}\left(\G_{S}^\top\y-\lambda \s_S\right)$$
Substituting the formula of $\tilde\thet_S$ into \eqref{eq:dL22} and denoting $\P_S:=\mathbf{I}-\G_{S}\left(\G_{S}^\top \G_{S}\right)^{-1} \G_{S}^\top$, we get
\begin{align*}
\s_{S^C}=\frac{1}{\lambda} \G_{S^C}^\top\P_S \y+\G_{S^C}^\top \G_{S}\left(\G_{S}^\top\G_{S}\right)^{-1} \s_S
\end{align*}
For $j\not\in S$, taking the $\ell_2$ norm and applying the triangular inequality give
\begin{align}
\|\s_{j}\|_2\leq\frac{1}{\lambda} \left\|\G_{j}^\top\P_S \y\right\|_2+\left\|\G_{j}^\top \G_{S}\left(\G_{S}^\top\G_{S}\right)^{-1} \s_S\right\|_2
\label{eq:sub j Cauchy}
\end{align}
Applying the Holder's inequality to the second term in \eqref{eq:sub j Cauchy} gives
\begin{align*}
&\left\|\G_{j}^\top \G_{S}\left(\G_{S}^\top\G_{S}\right)^{-1} \s_S\right\|_2
\leq\left\|\G_{j}^\top \G_{S}\left(\G_{S}^\top\G_{S}\right)^{-1}\right\|_2\left\|\s_S\right\|_\infty
<1-\gamma
\end{align*}
where the inequality follows from \Cref{assume:mutual incoherence} (mutual incoherence).

Regarding the first term in \eqref{eq:sub j Cauchy}, unlike in the LASSO support recovery analysis \cite{wainwright2009sharp} where the maximal inequality is directly applicable, we seek new tools since $\left\{\Big\|\G_{j}^\top\P_S \y\Big\|_2\right\}$ are non-centered random variables. We apply the Holder's inequality to the first term in \eqref{eq:sub j Cauchy},
\begin{align*}
\frac{1}{\lambda}\Big\|\G_{j}^\top\P_S \y\Big\|_2
\leq \frac{1}{\lambda}\Big\|\G_{j}^\top\Big\|_\infty\Big\|\P_S\Big\|_2 \Big\|\y\Big\|_\infty
\leq \frac{1}{\lambda}\Big\|\G_{j}^\top\Big\|_\infty \Big\|\y\Big\|_\infty
\end{align*}
in which the last inequality follows from the fact that $\P_S$ is a projection matrix with $\|\mathbf P_S\|_2\leq 1$.

All in all, we have
\begin{align*}
\max_{j\not\in S}\|\s_j\|_2\leq\frac{1}{\lambda}\max_{j\not\in S}\Big\|\G_{j}^\top\Big\|_\infty \Big\|\y\Big\|_\infty+1-\gamma
\end{align*}
and therefore, if $\lambda>\max\limits_{j\not\in S}\|\G_j^\top\|_\infty\|\y\|_\infty/\gamma$, then SNAM recovers the true support exactly. Notice that the matrix norm $\|\G_j^\top\|_\infty$ is the maximum of its $n$ absolute value column sums: $\|\G_j^\top\|_\infty=\max_{i=1}^n\big\|g_j([\X_j]_i,\w_j(0))\big\|_1$ where $g_j([\X_j]_i,\w_j(0))\in\R^m$.
\end{proof}

\subsection{Proofs in Section 5}
\begin{proof}[Proof of Theorem 5.1]
From \Cref{thm:slow_rate}, we see that $\frac{1}{n}\|f(\x)-h_n(\x)\|_2^2=O_p(1/\sqrt{n})=o_p(1)$. To prepare the proof of the convergence in probability measure, we consider the probaility space consisting of $(\mathcal{X}, E, \rho)$, where $\mathcal{X}$ is the sample space, $E$ is the event space, and $\rho$ is the probability measure. Defining the events $S_n:=\{x\in\mathcal{X}:|f(x)-h_n(x)|\geq \epsilon\}$, we have $S_n\in E$.

We will prove the theorem by contradiction. If there exists an $\epsilon>0$ such that for any $N, \delta>0$, there is some $n_N>N$ such that $\rho(\{x\in\mathcal{X}:|f(x)-h_n(x)|\geq \epsilon\})>\delta$. 

However, since 
\begin{align*}
\frac{1}{n}\|f(\x)-h_n(\x)\|_2^2    
&=\frac{1}{n}\sum_{i=1}^n(f(\x_i)-h_n(\x_i))_2^2\\
&\geq \frac{1}{n}\sum_{\x_i\in S_n}(f(\x_i)-h_n(\x_i))_2^2\\
&=\frac{1}{n}\sum_{i=1}^n \indict(\x_i\in S_n)(f(\x_i)-h_n(\x_i))_2^2\\
&\geq \frac{\epsilon^2}{n}\sum_{i=1}^n \indict(\x_i\in S_n)
\end{align*}
Denote each random variable $\indict(\x_i\in S_n):=Z_{n,i}$. Together they constitute a row-wise i.i.d. triangular array. Since $\sup_n \E(Z_{n,i}^2)\leq 1<\infty$, by applying the weak law of large number for triangular array \cite[Theorem 2.2.11]{durrett2019probability}, we obtain
\begin{align*}
&\frac{1}{n}\|f(\x)-h_n(\x)\|_2^2
\geq \frac{\epsilon^2}{n}\sum_{i=1}^n \indict(\x_i\in S_n)
\overset{p}{\to} \epsilon^2 \mathbb{P}(x\in S_n)>\epsilon^2\delta
\end{align*}
This contradicts with the asymptotic zero estimation MSE, i.e. $\frac{1}{n}\|f(\x)-h_n(\x)\|_2^2\overset{p}{\to}0$.
\end{proof}

\begin{proof}[Proof of Theorem 5.2 ]
Following the proof of \Cref{thm:consistency in measure}, we know for any $\epsilon>0, \delta>0$, there exists $N$ such that for any $n_N>N$, we have $\rho(\{x\in\mathcal{X}:|f(x)-h_n(x)|\geq \epsilon\})<\delta$ and denote $S_n(\epsilon):=\{x\in\mathcal{X}:|f(x)-h_n(x)|\geq \epsilon\}$. We further denote $S_{n,j}^C:=\{x_{-j}:(x_j,x_{-j})\in S_n^C\}$ where $S_n^C$ is the complement of $S_n$.

Under the condition that $\mathcal{X}_{j}$ is independent of $\mathcal{X}_{-j}$, we take the expectation with respect to $\mathcal{X}_{-j}$, using the marginal density as $p_{-j}$:
\begin{align*}
&\int_{S_{n,j}^C} f(\mathcal{X})p_{-j}(u)du
=\int_{S_{n,j}^C} (f_{n,j}(\mathcal{X}_j)+f_{n,-j}(u))p_{-j}(u)du
=f_{n,j}(\mathcal{X}_j)+c_{j,1}
\end{align*}

Notice that this integral is also bounded between  $\int_{S_{n,j}^C}(h_{n,j}(\mathcal{X}_j)+h_{n, -j}(u)\pm\epsilon)p_{-j}(u)du=\mathbb{P}(\mathcal{X}_{-j}\in S_{n,j}^C)(h_{n,j}(\mathcal{X}_j)\pm\epsilon)+c_{j,2}$. The probability $\mathbb{P}(\mathcal{X}_{-j}\in S_{n,j}^C)$ goes to 1 as $\delta\to 0$. Further, as $\epsilon\to 0$, we have $h_{n,j}(\mathcal{X}_j)\overset{p}{\to}f(\mathcal{X}_j)+c_j$ for some constant $c_j$.
\end{proof}

\section{Training SNAM}
\subsection{Different optimizers}
\label{app:train SNAM}
Here we present the detailed optimizers to train the SNAM. Denoting all trainable parameters in all layers of the $j$-th sub-network as $\Theta_j$, and the loss as $\mathcal{L}+\lambda\sum_j\|\Theta_j\|_2$, then we have
\begin{enumerate}
    \item \textbf{Subgradient method:}
    \begin{align*}
    &\Theta_j(t+1)
    =\Theta_j(t)
    -\eta\left(\frac{\partial \mathcal{L}}{\partial \Theta_j(t)}+\lambda\frac{\Theta_j(t)}{\|\Theta_j(t)\|_2}\cdot 1\{\Theta_j(t)\neq\mathbf{0}\}\right)
    \end{align*}

    \item \textbf{Proximal gradient descent:}
    $$\Theta_j(t+1)=\text{Prox}_{\lambda\eta}\left(\Theta_j(t)-\eta\frac{\partial \mathcal{L}}{\Theta_j(t)}\right)$$
    \begin{align*}
    \text{where } \text{Prox}_\gamma(x)=
    \begin{cases}
    x-\gamma\frac{x}{\|x\|_2}, &x>\gamma
    \\
    0, &x\leq \gamma
    \end{cases}
    \end{align*}
\end{enumerate}

Notice that for subgradient methods, we can use Adam, Adagrad, momentum and so on, as long as we set the subgradient of zero vector to be zero. For proximal gradient methods, we can use momentums as well, e.g. in FISTA \cite{beck2009fast}.

\subsection{Different optimization problems}
\label{app:different SNAM}
Here we introduce more SNAMs that can perform feature selection using different group penalties.
\begin{enumerate}
    \item \textbf{SNAM extended from LASSO (the regular one):}
    $$\min_{\Theta_j}\mathcal{L}+\lambda\sum_j \|\Theta_j\|_2$$
    \item \textbf{SNAM extended from SLOPE:}
    $$\min_{\Theta_j}\mathcal{L}+\sum_j \lambda_j(\{\|\Theta_k\|_2\})_{(j)}$$
    where $\bm\lambda\in\R^p$ with $\lambda_j>\lambda_{j+1}$, and $(\bm v)_{(j)}$ is the $j$-th largest element in the vector $\bm v$. I.e. the largest norm is penalized with heaviest penalty.
    \item \textbf{SNAM extended from 2-level SLOPE:}
    $$\min_{\Theta_j}\mathcal{L}+\lambda_1\sum_{j\leq m} (\{\|\Theta_k\|_2\})_{(j)}+\lambda_2\sum_{m<j\leq p} (\{\|\Theta_k\|_2\})_{(j)}$$
    where $\bm\lambda$ contains $m $ elements as $\lambda_1$ and $p-m$ elements as $\lambda_2$.
    \item \textbf{SNAM extended from adaptive LASSO:}
    $$\min_{\Theta_j}\mathcal{L}+\lambda\sum_j w_j\|\Theta_j\|_2$$
    where $w_j\in\R$ is the weight to adjust LASSO's bias. Some suggestions are $w_j=1/\|\Theta_{j,\text{NAM}}\|_2$ or $1/\|\Theta_{j,\text{SNAM}}\|_2$.
    \item \textbf{SNAM extended from elastic net:}
    $$\min_{\Theta_j}\mathcal{L}+\lambda_1\sum_j \|\Theta_j\|_2+\lambda_2\sum_j \|\Theta_j\|_2^2$$
    Note this is regular SNAM with weight decay.
\end{enumerate}

\section{Figure Zoo}
\label{app:fig zoo}
\begin{figure}[!htb]
    \centering
    \includegraphics[width=\linewidth]{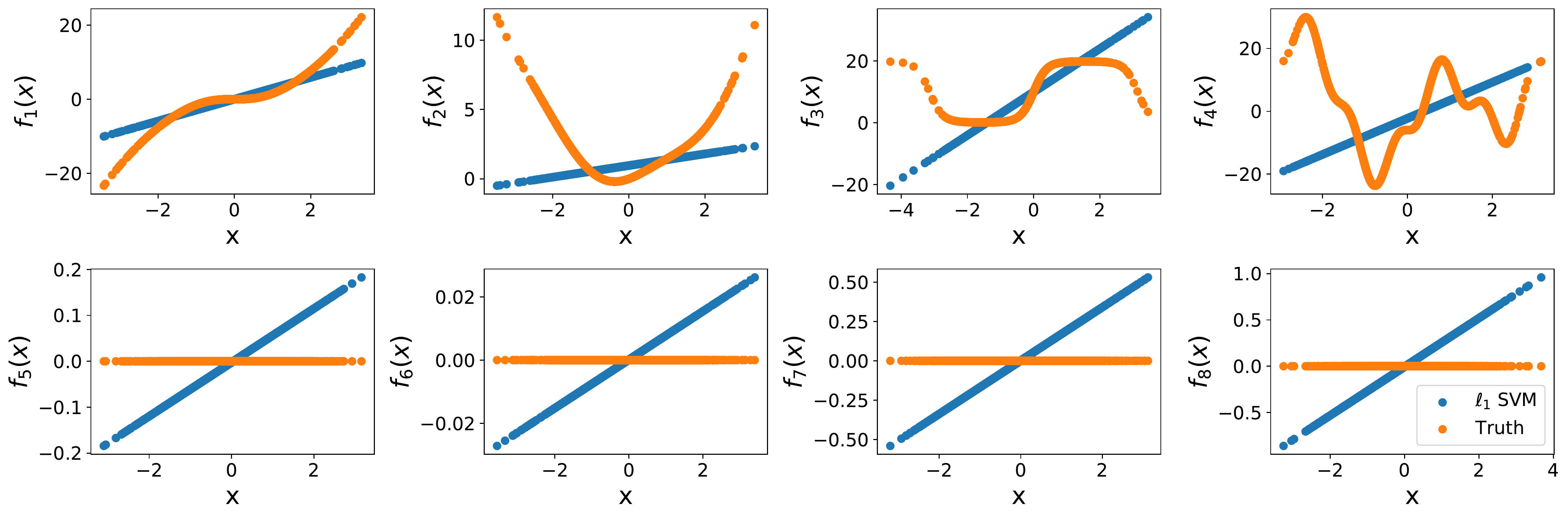}
    \caption{Individual effect learned by $\ell_1$ SVM on synthetic regression. Blue dots are prediction $\hat f_j(\X_j)$ and orange dots are truth $f_j(\X_j)$, with $j=1,\cdots,8$.}
    \label{fig:regression_l1svm}
\end{figure}

\begin{figure}[!htb]
    \centering
    \includegraphics[width=\linewidth]{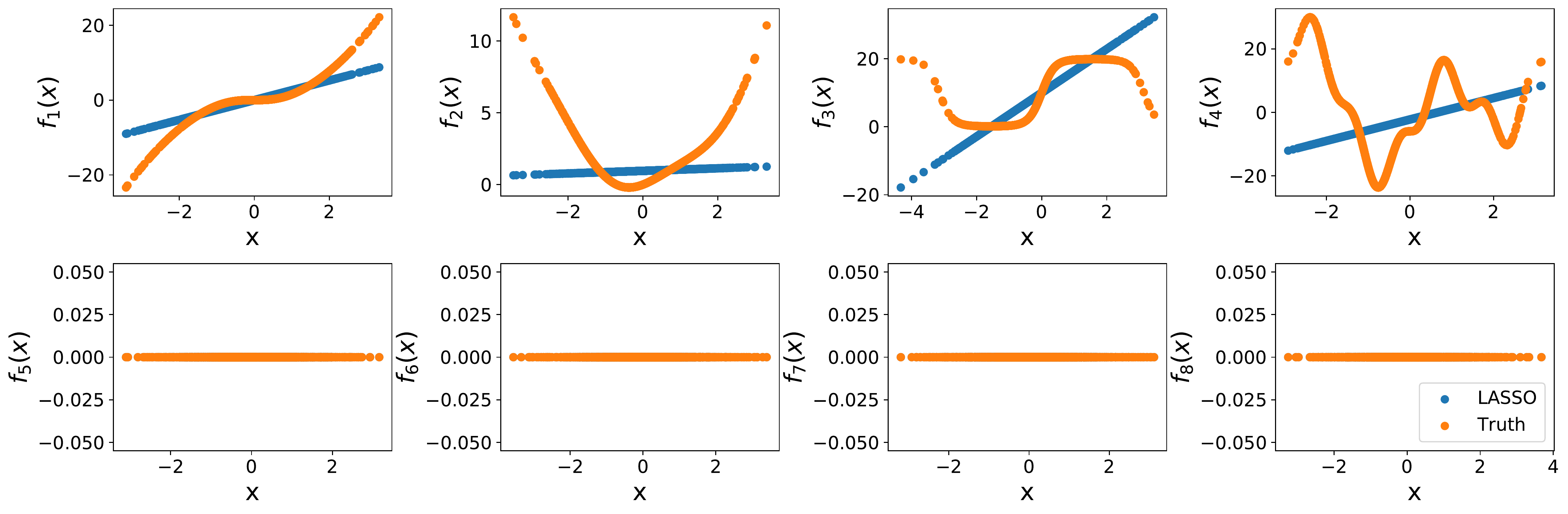}
    \caption{Individual effect learned by LASSO on synthetic regression. Blue dots are prediction $\hat f_j(\X_j)$ and orange dots are truth $f_j(\X_j)$, with $j=1,\cdots,8$.}
    \label{fig:regression_lasso}
\end{figure}

\begin{figure}[!htb]
    \centering
    \includegraphics[width=\linewidth]{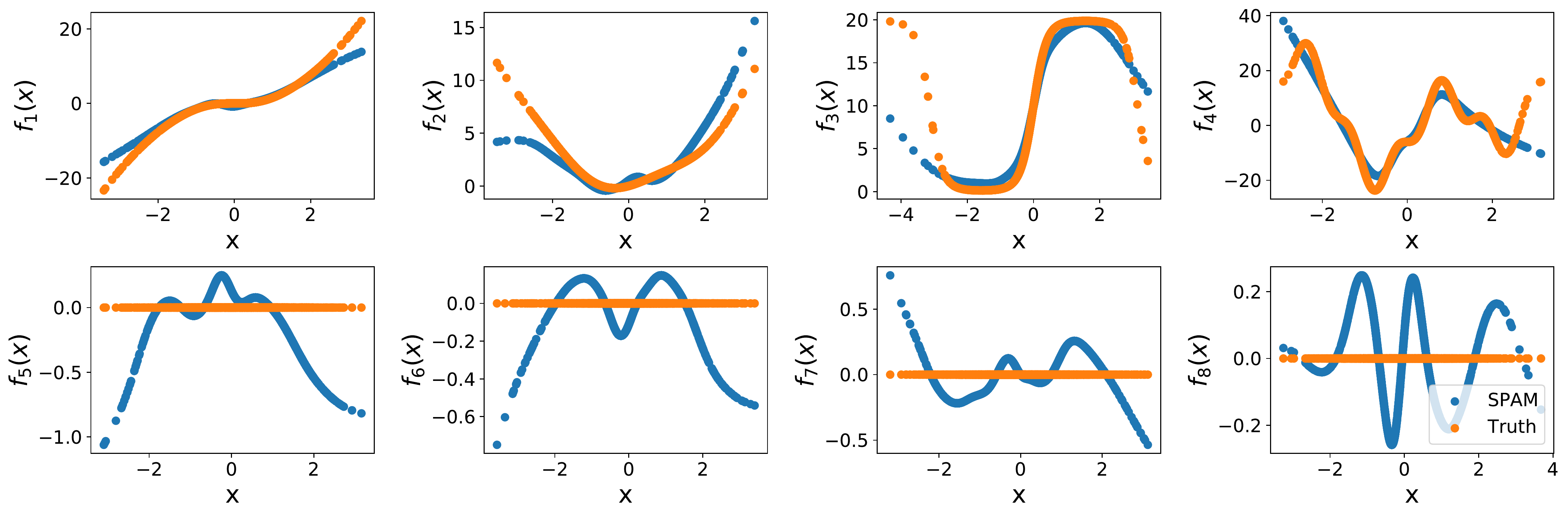}
    \caption{Individual effect learned by SPAM on synthetic regression. Blue dots are prediction $\hat f_j(\X_j)$ and orange dots are truth $f_j(\X_j)$, with $j=1,\cdots,8$.}
    \label{fig:regression_spam}
\end{figure}

In this section, we show some additional experiments to further verify our conclusions. In \Cref{fig:regression_l1svm},\Cref{fig:regression_lasso}, and \Cref{fig:regression_spam}, we show the approximation of each methods to the true function. As we mentioned in the main text, all the functions are zero function except the top 4. Since the top 4 functions are non-linear, there is no surprise that $\ell_1$ SVM and LASSO have bad performance. For SPAM, although the approximation is better than $\ell_1 SVM$ and $LASSO$, it is still beaten by our SNAM, especially in jumpy functions like $f_4(\x)$. The models are well tuned and the dataset is the same as in \Cref{fig:regression_snam}. We only show the top 8 figures for simplicity.
The individual effect of different physical quantities to predict the critical temperature of super-conductors is demonstrated as below. Features that have small values can be taken as insignificant factors. 
\begin{figure}[!htb]
    \centering
    \includegraphics[width=\linewidth]{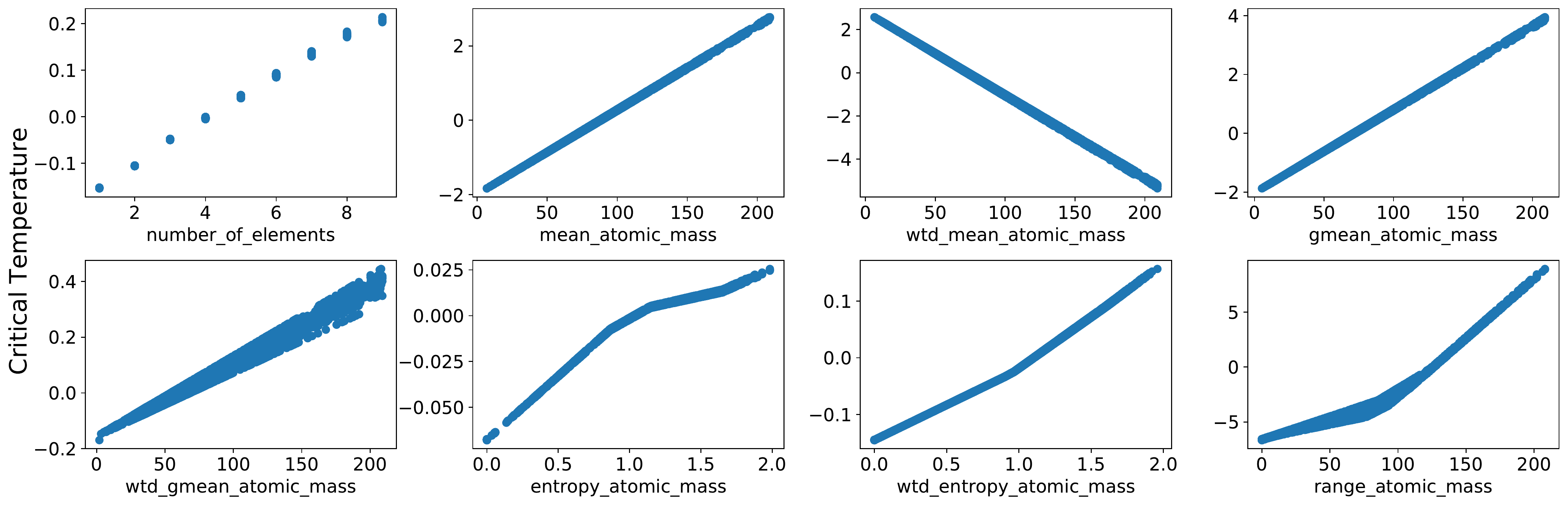}
    \caption{Individual effect learned by SNAM on the super-conductivity dataset.}
    \label{fig:super_spam}
\end{figure}

\section{Experiment Details} 
\label{app:experiment detail}
For all the experiments except the one with super-conductivity dataset, we apply the same three-layer architecture for the sub-networks, using ReLU activation. The neurons in the first, the second hidden layers and the output layer are 100, 50, 1 respectively for regression tasks and 100, 50, 2 for binary classification tasks. For the super-conductivity dataset, the sub-networks are two-layer neural networks with 16 hidden neurons for each. Notice that sub-networks have bias terms in hidden layers but not for output layers, since there is a global bias $\beta$ to be added to the outputs.

The optimizer is Adam by default with batch size 256, except for the super-conductivity dataset, the batch size is 512.

The hyperparameters of SNAM are listed below.
\begin{table}[!ht]
	\centering
	\begin{tabular}[t]{ccccc}
		\hline
		Hyperparameters&Learning rate&Penalty& Epoch \\
        \hline
		Synthetic regression&$5\times 10^{-3}$&2&100&\\
		\hline
		Synthetic classification&$5\times 10^{-3}$&0.04&20&\\
		\hline
		California Housing&$ 10^{-5}$&1&300&\\
		\hline
		COMPAS&$5\times 10^{-3}$&0.08&100&\\
		\hline
		Super-conductivity&$5\times 10^{-3}$&10&20&\\
\hline
	\end{tabular}
    \caption{Hyperparameters of experiments.}
    \label{tab:hyper}
\end{table}

In the experiments over optimizers in \Cref{fig:optimizer}, we apply the same hyperparametes for Adam, SGD and gradient descent: learning rate is $10^{-6}$, penalty is $1$ and the number of epoch is $300$.

\subsection{Miscellaneous}

\begin{algorithm}[!htbp]
\caption{SPAM Backfitting Algorithm}
\label{alg:SPAM}
\begin{algorithmic}
\STATE{\textbf{Input}} Data \( \left(\X_{i}, y_{i}\right) \), regularization parameter \( \lambda \).\\
\STATE{\textbf{Initialize}} \( \widehat{f}_{j}=0 \), for \( j=1, \ldots, p \)\\
\STATE{\textbf{Do} until \( \hat{f}_{j} \) converge:}
\STATE{\textbf{\quad For}} each \( j=1, \ldots, p \) :
\STATE{\textbf{\quad\quad (a)}} Compute the residual: \( R_{j}=\y-\sum_{k \neq j} \widehat{f}_{k}\left(\X_{k}\right) \)\\
\STATE{\textbf{\quad\quad (b)}} Estimate \( P_{j}=\mathbb{E}\left[R_{j} \mid \X_{j}\right] \) by smoothing: \( \widehat{P}_{j}=\mathcal{S}_{j} R_{j} \)\\
\STATE{\textbf{\quad\quad (c)}} Estimate norm: \( \widehat{s}_{j}^{2}=\frac{1}{n} \sum_{i=1}^{n} \widehat{P}_{j}^{2}(\X_{ij}) \)\\
\STATE{\textbf{\quad\quad (d)}} Soft-threshold: \( \widehat{f}_{j}=\left[1-\lambda / \widehat{s}_{j}\right]_{+} \widehat{P}_{j} \)\\
\STATE{\textbf{\quad\quad (e)}}
 Center: \( \widehat{f}_{j} \leftarrow \widehat{f}_{j}-\sum_{i=1}^{n} \hat{f}_{j}\left(\X_{i j}\right)/n \)\\
\STATE{\textbf{Output}} Individual functions \( \widehat{f}_{j} \)
\end{algorithmic}
\end{algorithm}



\end{document}